\pdfoutput=1

\documentclass[11pt]{article}

\usepackage[]{naacl2021}

\usepackage{times}
\usepackage{latexsym}

\usepackage[T1]{fontenc}

\usepackage[utf8]{inputenc}

\usepackage{microtype}
\usepackage{listings}
\usepackage{booktabs}
\usepackage{graphicx}
\usepackage{makecell}
\usepackage{subcaption}
\usepackage{enumerate}
\usepackage[ruled,vlined]{algorithm2e}
\usepackage{lipsum}
\newcommand\blfootnote[1]{%
  \begingroup
  \renewcommand\thefootnote{}\footnote{#1}%
  \addtocounter{footnote}{-1}%
  \endgroup
}

\DeclareCaptionFormat{mylst}{\hrule#1#2#3}
\title{MUDES: Multilingual Detection of Offensive Spans
}

\author{Tharindu Ranasinghe \\
  University of Wolverhampton \\
  Wolverhampton, UK\\
  \texttt{tharindu.ranasinghe@wlv.ac.uk} \\\And
  Marcos Zampieri \\
  Rochester Institute of Technology \\
  Rochester, NY, USA \\
  \texttt{mazgla@rit.edu} \\}

\begin{document}
\maketitle
\begin{abstract}

The interest in offensive content identification in social media has grown substantially in recent years. Previous work has dealt mostly with post level annotations. However, identifying offensive spans is useful in many ways. To help coping with this important challenge, we present MUDES, a multilingual system to detect offensive spans in texts. MUDES features pre-trained models, a Python API for developers, and a user-friendly web-based interface. A detailed description of MUDES' components is presented in this paper.

\end{abstract}

\section{Introduction}

Offensive and impolite language are widespread in social media posts motivating a number of studies on automatically detecting the various types of offensive content (e.g. aggression \cite{kumar2018benchmarking,trac2020}, cyber-bullying \cite{rosa2019automatic}, hate speech \cite{malmasi2018}, etc.). 
Most previous work has focused on classifying full instances (e.g. posts, comments, documents) (e.g. offensive vs. not offensive) while the identification of the particular spans that make a text offensive has been mostly neglected. \blfootnote{WARNING: This paper contains text excerpts and words that are offensive in nature.}

Identifying offensive spans in texts is the goal of the SemEval-2021 Task 5: Toxic Spans Detection \cite{pav2020semeval}. The organisers of this task argue that highlighting toxic spans in texts helps assisting human moderators (e.g. news portals moderators) and that this can be a first step in semi-automated content moderation. Finally, as we demonstrate in this paper, addressing offensive spans in texts will make the output of offensive language detection systems more interpretable thus allowing a more detailed linguistics analysis of predictions and improving the quality of such systems. 

With these important points in mind, we developed MUDES: {\bf Mu}ltilingual {\bf De}tection of Offensive {\bf S}pans. MUDES is a multilingual framework for offensive language detection focusing on text spans. The main contributions of this paper are the following:

\begin{enumerate}
\item We introduce MUDES, a new Python-based framework to identify offensive spans with state-of-the-art performance.
\item We release four pre-trained offensive language identification models: en-base, en-large models which are capable of identifying offensive spans in English text. We also release Multilingual-base and  Multilingual-large models which are able to recognise offensive spans in languages other than English.
\item We release a Python Application Programming Interface (API) for developers who are interested in training more models and performing inference in the code level. 
\item For general users and non-programmers, we release a user-friendly web-based User Interface (UI), which provides the functionality to input a text in multiple languages and to identify the offensive span in that text.
\end{enumerate}

\section{Related Work}
\label{sec:RW}

\begin{table*}[t]
\centering
\begin{tabular}{p{9cm}p{6cm}}
\toprule
\bf Post & \bf Offensive Spans  \\
\midrule
 \textcolor{red}{Stupid} hatcheries have completely \textcolor{red}{fucked} everything & [0, 1, 2, 3, 4, 5, 34, 35, 36, 37, 38, 39]  \\
 Victimitis: You are such an \textcolor{red}{asshole}. & [28, 29, 30, 31, 32, 33, 34]  \\
 So is his mother. They are silver spoon parasites. & []  \\
 You're just \textcolor{red}{silly}. & [12, 13, 14, 15, 16]  \\
\bottomrule
\end{tabular}
\caption{Four comments from the dataset, with their annotations. The offensive words are displayed in red and the spans are indicated by the character position in the instance.}
\label{tab:examples}
\end{table*}

Early approaches to offensive language identification relied on traditional machine learning classifiers \cite{dadvar2013improving} and later on neural networks combined with word embeddings \cite{majumder2018filtering, hettiarachchi-ranasinghe-2019-emoji}. Transformer-based models like BERT \cite{devlin2019bert} and ELMO \cite{peters-etal-2018-deep} have been recently applied to offensive language detection achieving competitive scores \cite{wang-etal-2020-galileo, ranasinghe-hettiarachchi-2020-brums} in recent SemEval competitions such as HatEval \cite{basile2019semeval} OffensEval \cite{zampieri-etal-2020-semeval}. 

In terms of languages, the majority of studies on this topic deal with English \cite{malmasi2017detecting,yao2019cyberbullying,ridenhour2020detecting,rosenthal2020} due to the the wide availability of language resources such as corpora and pre-trained models. In recent years, several studies have been published on identifying offensive content in other languages such as Arabic \cite{mubarak2020arabic}, Dutch \cite{tulkens2016dictionary}, French \cite{chiril-etal-2019-multilingual}, Greek \cite{pitenis2020}, Italian \cite{poletto2017hate}, Portuguese \cite{fortuna2019hierarchically}, and Turkish \cite{coltekin2020}. Most of these studies have created new datasets and resources for these languages opening avenues for multilingual models as those presented in \newcite{ranasinghe-etal-2020-multilingual}. However, all studies presented in this section focused on classifying full texts, as discussed in the Introduction. MUDES' objective is to fill this gap and perform span level offensive language identification.



\section{Data}
The main dataset used to train the machine learning models presented in this paper is the dataset released within the scope of the aforementioned SemEval-2021 Task 5: Toxic Spans Detection for English.
The dataset contains posts (comments) from the publicly available Civil Comments dataset \cite{borkan2019nuanced}. The organisers have randomly selected 10,000 posts, out of a total of 1,2 million posts in the original dataset. The offensive spans have been annotated using a crowd-annotation platform, employing three crowd-raters per post. By the time of writing this paper, only the trial set and the training set have been released and the gold labels for the test set have not yet been released. Therefore, training of the machine learning models presented in MUDES was done on the training set which we refer to as \textit{TSDTrain} and the evaluation was conducted on the trial set which we refer to as \textit{TSDTrial} set. In Table \ref{tab:examples} we show four randomly selected examples from the \textit{TSDTrain} dataset with their annotations.

The general idea is to learn a robust model from this dataset and generalize to other English datasets which do not contain span annotation. Another goal is to investigate the feasibility of annotation projection to other languages.

\begin{figure*}[ht]
\centering
\includegraphics[scale=0.55]{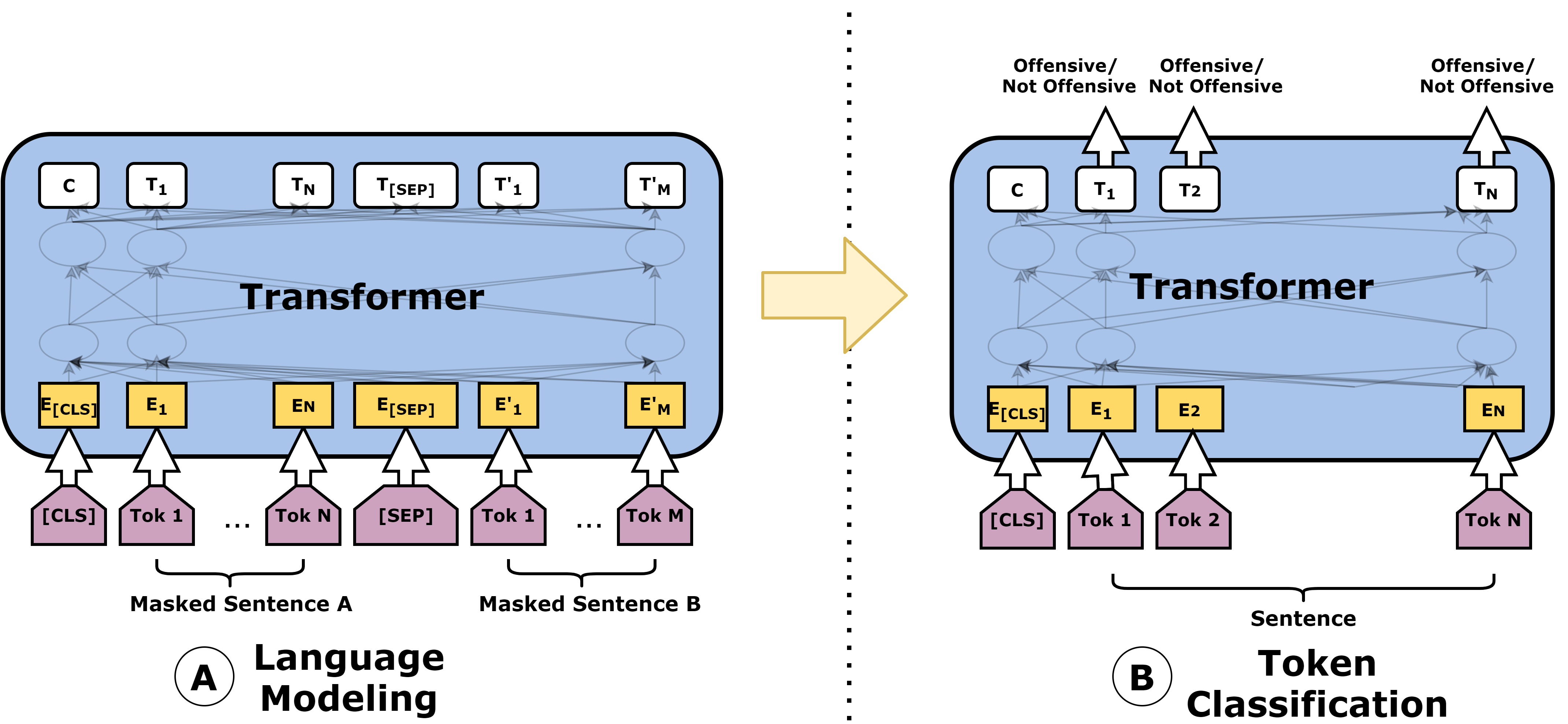}
\caption{Model Architecture. Architecture consists of two parts. Part A is the language modelling and Part B is the token classification.}
\label{fig:architecture}
\end{figure*}

\paragraph{Other Datasets} In order to evaluate our framework in different domains and languages we used three publicly available offensive language identification datasets. As an off-domain English dataset, we choose the Offensive Language Identification Dataset (OLID) \cite{OLID}, used in OffensEval 2019 (SemEval-2019 Task 6) \cite{offenseval}, containing over 14,000 posts from Twitter. To evaluate our framework in different languages, we selected a Danish \cite{sigurbergsson2020offensive} and a Greek \cite{pitenis2020} dataset. These two datasets have been provided by the organisers of OffensEval 2020 (SemEval-2020 Task 12) \cite{zampieri-etal-2020-semeval} and were annotated using OLID's annotation guidelines. The Danish dataset contains over 3,000 posts from Facebook and Reddit while the Greek dataset contains over 10,000 Twitter posts, allowing us to evaluate our dataset in an off-domain, multilingual setting. As these three datasets have been annotated at the instance level, we followed an evaluation process explained in Section \ref{sec:results}.



\section{Methodology}
The main motivation behind this methodology is the recent success that transformer models had in various NLP tasks \cite{devlin2019bert} including offensive language identification  \cite{ranasinghe-etal-2020-multilingual, ranasinghe2019brums,wiedemann-etal-2020-uhh}. Most of these transformer-based approaches take the final hidden state of the first token ([CLS]) from the transformer as the representation of the whole sequence and a simple softmax classifier is added to the top of the transformer model to predict the probability of a class label \cite{10.1007/978-3-030-32381-3_16}. However, as previously mentioned, these models classify whole comments or documents and do not identify the spans that make a text offensive. Since the objective of this task is to identify offensive spans rather than classifying the whole comment, we followed a different architecture.

As shown in Figure \ref{fig:architecture}, the complete architecture contains two main parts; Language Modeling (LM) and Token Classification (TC). In the LM part, we used a pre-trained transformer model and retrained it on the \textit{TSDTrain} dataset using Masked Language Modeling (MLM). In the second part of the architecture, we used the saved model from the LM part and we perform a token classification. We added a token level classifier on top of the transformer model as shown in Figure \ref{fig:architecture}. The token-level classifier is a linear layer that takes the last hidden state of the sequence as the input and produce a label for each token as the output. In this case each token can have two labels; offensive and not offensive. We have listed the training configurations in the Appendix.

We experimented with several popular transformer models like BERT \cite{devlin2019bert}, XLNET \cite{NEURIPS2019_dc6a7e65}, ALBERT \cite{Lan2020ALBERT}, RoBERTa \cite{liu2019roberta} etc. From the pre-trained transformer models we selected, we grouped the large models and base models separately in order to release two English models. A large model; en-large which is more accurate, but has a low efficiency regarding space and time. The base model; en-base is efficient, but has a comparatively low accuracy than the en-large model. All the experiments have been executed for five times with different random seeds and we took the mode of the classes predicted by each random seed as the final result \cite{hettiarachchi-ranasinghe-2020-infominer}.

\paragraph{Multilingual models} - The motivation behind the use of multilingual models comes from recent works \cite{ranasinghe-etal-2020-multilingual, ranasinghetallip, ranasinghe2020wlv} which used transfer learning and cross-lingual embeddings. These studies show that cross-lingual transformers like XLM-R \cite{conneau2019unsupervised} can be trained on an English dataset and have the model weights saved to detect offensive language in other languages outperforming monolingual models trained on the target language dataset. We used a similar methodology but for the token classification architecture instead. We used XLM-R cross-lingual transformer model \cite{conneau2019unsupervised} as the Transformer in Figure \ref{fig:architecture} on \textit{TSDTrain} and carried out evaluations on the Danish and Greek datasets. We release two multilingual models; multilingual-base based on XLM-R base model and multilingual-large based on XLM-R large model.

\section{Evaluation and Results}
\label{sec:results}
We followed two different evaluation methods. In Section \ref{sec:tsd_eval} we present the methods used to evaluate offensive spans on the TSDTrial set. In Section \ref{subsec:other_eval} we presented the methods used to evaluate the other three datasets which only contained post level annotations. 

\subsection{Offensive Spans Evaluation}
\label{sec:tsd_eval}

For the Toxic Spans Detection dataset, we followed the same evaluation procedure of the SemEval Toxic Spans Detection competition. The organisers have used F1 score mentioned in \citet{da-san-martino-etal-2019-fine} to evaluate the systems. Let system $A_i$ return a set $S^t_{A_i}$ of character offsets, for parts of the post found to be toxic. Let $G_t$ be the character offsets of the ground truth annotations of t. We compute the F1 score of system $A_i$ with respect to the ground truth G for post t as mentioned in Equation \ref{equation_f1}, where $\vert$ ·$\vert$ denotes set cardinality.


\begin{equation}
\label{equation_f1}
F_{1}^{t}\left(A_{i}, G\right)=\frac{2 \cdot P^{t}\left(A_{i}, G\right) \cdot R^{t}\left(A_{i}, G\right)}{P^{t}\left(A_{i}, G\right)+R^{t}\left(A_{i}, G\right)}
\end{equation}

$P^{t}\left(A_{i}, G\right)=\frac{\left|S_{A_{i}}^{t} \cap S_{G}^{t}\right|}{\left|S_{A_{i}}^{t}\right|}$
$R^{t}\left(A_{i}, G\right)=\frac{\left|S_{A_{i}}^{t} \cap S_{G}^{t}\right|}{\left|S_{G}^{t}\right|}$

\vspace{7mm}

\noindent We present the results along with the baseline provided by the organisers in Table \ref{tab:results}. The baseline is implemented using a spaCy NER pipeline. The spaCy NER system contains a word embedding strategy using sub word features and Bloom embedding \cite{10.1145/3109859.3109876}, and a deep convolution neural network with residual connections. Additionally, we compare our results to a lexicon-based word match approach mentioned in \citet{ranasinghe2021semeval} where the lexicon is based on profanity words from online resources\footnote{\url{https://www.cs.cmu.edu/~biglou/resources/bad-words.txt}}\textsuperscript{,}\footnote{\url{https://github.com/RobertJGabriel/Google-profanity-words}}.

\begin{table}[!ht]
\begin{center}
\scalebox{0.83}{
\begin{tabular}{ c|c|c } 
 \hline
 \textbf{Model Name} & \textbf{Base Model} & \textbf{F1 score} \\ 
 \hline
   en-large & roberta-large & 0.6886 \\
   en-base & xlnet-base-cased & 0.6734 \\
  \hline
   multilingual-large & XLM-R-large & 0.6338 \\
   multilingual-base & XLM-R-base & 0.6160 \\
   \hline
   spaCy baseline & NA & 0.5976 \\
   \hline
   \makecell{Lexicon word match \\ \cite{ranasinghe2021semeval}}  & NA & 0.3378 \\
   \hline
\end{tabular}
}
\end{center}
\caption{Results ordered by F1 score for TSD Trial.} 
\label{tab:results}
\end{table}

\noindent The results show that all MUDES' models outperform the spaCy baseline and the lexicon-based word match. From all of the large transformer models we experimented roberta-large performed better than others. Therefore, we released it as en-large model in MUDES. From the base models we experimented, XLNet-base-cased model outperformed all the other base models so we released it as en-base model. We also released two multilingual models; multilingual-base and multilingual-large based on XLM-R-base and XLM-R-large respectively. All the pre-trained MUDES' models are available to download from HuggingFace model hub \footnote{MUDES' models are available on \url{https://huggingface.co/mudes}} \cite{wolf-etal-2020-transformers}.

\subsection{Off-Domain and Multilingual Evaluation}
\label{subsec:other_eval}

For the English off-domain and multilingual datasets we followed a different evaluation process. We used a pre-trained MUDES' model trained on \textit{TSDTrain} to predict the offensive spans for all texts in the test sets of two non-English datasets (Danish, and Greek) and English off-domain dataset, OLID, which is annotated at the document level. If a certain text contains at least one offensive span we marked the whole text as offensive following the OLID annotation guidelines described in \newcite{OLID}. We compared our results to the best systems submitted to OffensEval 2020 in terms of macro F1 reported by the task organisers \cite{zampieri-etal-2020-semeval}. We present the results along with the majority class baseline for each dataset in Table \ref{tab:all}. For English off domain dataset (OLID) we only used the MUDES en models while for Danish and Greek datasets we used the MUDES multilingual models.

\begin{table}[!ht]
\centering
\setlength{\tabcolsep}{4.5pt}
\scalebox{0.85}{
\begin{tabular}{llcc}
\hline
\bf Language & \bf Model   & \bf M F1 &  \\ \hline

& \newcite{pamies-etal-2020-lt} & 0.8119  \\
Danish & multilingual-large & 0.7623         \\
& multilingual-base       & 0.7143         \\
& Majority Baseline & 0.4668 &	 \\
 \hline

 & \newcite{wiedemann-etal-2020-uhh} &  0.9204  \\
English & en-large       & 0.9023         \\
& en-base       & 0.8892         \\
& Majority Baseline & 0.4193  \\
 \hline
 
& \newcite{ahn-etal-2020-nlpdove} & 0.8522 \\
Greek & multilingual-large       & 0.8143         \\
& multilingual-base      & 0.7820         \\
& Majority Baseline & 0.4202 	 \\
\hline
\end{tabular}
}
\caption{Results ordered by macro (M) F1 for Danish, English and Greek datasets}
\label{tab:all}
\end{table}

\noindent Results show that despite the change of domain and the language, MUDES perform well in all the datasets and compares favourably to the best systems submitted. It should be noted that the best systems have been predominantly trained on offensive languages identification task on post level while MUDES' objective is different. Yet MUDES come closer to the best systems in all the datasets. 

From the results, it is clear that MUDES english models can perform in a different domain like Twitter. Also the results show that MUDES multilingual models are capable of identifying offensive spans in other languages too. Since XLM-R supports 104 languages, this approach will benefit all those languages without any training data at all.

\section{System Demonstration}
\label{sec:demo}

\subsection{Application Programming Interface}
\label{sec:api}
MUDES is available as a Python package in the Python Package Index (PyPI)\footnote{\url{https://pypi.org/project/mudes/}}. The package is related to MUDES GitHub repository\footnote{\url{https://github.com/tharindudr/MUDES}}. Users can install it easily with the following command after installing PyTorch \cite{NEURIPS2019_9015}.

\begin{lstlisting}[basicstyle=\small, language=bash]
  $ pip install mudes
\end{lstlisting}

\noindent The Python package contains the following functionalities. 



\begin{figure*}
\centering
  \begin{subfigure}[b]{7cm}
    \centering\includegraphics[width=7cm]{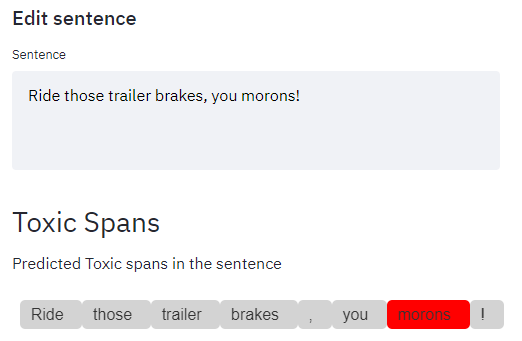}
    \caption{Example from Civil Comments Dataset}
    \label{fig:civil_results}
  \end{subfigure}
  \begin{subfigure}[b]{7cm}
    \centering\includegraphics[width=7cm]{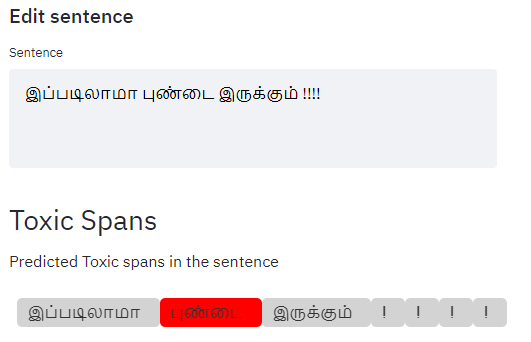}
    \caption{Example from Tamil. (\textcolor{red}{Pussy} like this !!!!)}
    \label{fig:tamil_Results}
  \end{subfigure}
\caption{Examples in English and in a low-resource languages. The experiments were conducted with en-large and the multilingual-large models respectively.}
\label{fig:low_results}
\end{figure*}

\paragraph{Get offensive spans with a pretrained model} The library provides the functionality to load a pretrained model and use it to identify offensive spans. The following code segment downloads and loads MUDES' en-base model in a CPU only environment and identifies offensive spans in the the text; \textit{"This is fucking crazy!!"}. If the users prefer a GPU, the argument \textit{use\_cuda} should be set to True.

\begin{lstlisting}[caption={English Inference Example}]
from mudes.app.mudes_app 
                import MUDESApp
  
sentence = "This is fucking crazy!!"
  
app = MUDESApp("en-base", 
                use_cuda=False)
app.predict_toxic_spans(sentence)
\end{lstlisting}

\paragraph{Train a MUDES model} The library provides the functionality to train a MUDES model from scratch using the code segment present next. It takes a Pandas dataframe in the format of {\em TSDTrain}, formats it for the token classification task and train a MUDES model from scratch. MUDES support popular transformer types as bert, xlnet, roberta etc. as the MODEL\_TYPE and name of the model as appear in Hugging Face \cite{wolf-etal-2020-transformers} model repository. \footnote{\url{https://huggingface.co/models}}

\begin{lstlisting}[caption={Training Example}]
from mudes.algo.mudes_model 
                import MUDESModel
from mudes.algo.preprocess 
                import read_datafile, 
                       format_data

train_df = format_data(train)
tags = train_df['labels']
            .unique().tolist()
            
model = MUDESModel(MODEL_TYPE, 
            MODEL_NAME, labels=tags)
model.train(train_df)
\end{lstlisting}


\subsection{User Interface}
\label{sec:ui}
We developed a prototype of the User Interface (UI) to demonstrate the capabilities of the system. The UI is based on Streamlit\footnote{\url{www.streamlit.io}} which provides functionalities to easily develop dashboards for machine learning projects. The code base for the UI is available in GitHub \footnote{\url{https://github.com/tharindudr/MUDES-UI}.}. This UI is hosted in a Linux server. \footnote{\url{http://rgcl.wlv.ac.uk/mudes/}} We also release a Docker container image of the UI in Docker Hub\footnote{Docker Hub is a hosted repository service provided by Docker for finding and sharing container images.} for those who are interested in self hosting the UI. Docker enables developers to easily deploy and run any application as a lightweight, portable, self-sufficient container, which can run virtually anywhere. The released Docker container image follows Continuous Integration/Continuous Deployment (CI/CD) from the GitHub repository which allows sharing and deploying the code quickly and efficiently.

Once Docker is installed, one can easily run our UI with this command. 

\begin{lstlisting}[basicstyle=\small, language=bash]
  $ docker run tharindudr/mudes
\end{lstlisting}

\noindent This command will automatically install all the required packages, download and load the pre-trained models and open the system in the default browser. We provide the following functionalities from the user interface. 

\paragraph{Switch through pretrained models} - The users can switch through the pre-trained models using the radio buttons available in the left side of the UI under Available Models section. They can select an option from en-base, en-large, multilingual-base and multilingual-large. These models have been already downloaded from the HuggingFace model hub and they are loaded in to the random-access memory of the hosting computer.

\paragraph{Switch through available datasets} - We have made the four datsets used in this paper available from the UI for the users to experiment with \cite{borkan2019nuanced,OLID,pitenis2020,sigurbergsson2020offensive}. Once the user selects a particular option, the system will automatically load the test set of the selected dataset. Once it is loaded the user can iterate through the dataset using the scrollbar. For each text the UI will display the offensive spans in red.

\paragraph{Get offensive spans for a custom text} - The users can also enter a custom text in the text box, hit ctrl+enter and see the offensive spans available in the input text. Once processed through the system, any offensive spans available in the text will be displayed in red. Figure \ref{fig:low_results} shows several screenshots from the UI. It illustrates an example on English for the texts taken from civil comments dataset \cite{borkan2019nuanced} conducted with en-large model. To show that MUDES framework works on low resource language too, Figure \ref{fig:low_results} also displays an example from Tamil.



\subsection{System Efficiency}
\label{sec:eff}
The time taken to predict the offensive spans for a text will be critical in an online system developed for real time use. Therefore, we evaluated the time MUDES takes to predict the offensive spans in 100 texts for all the released models in a CPU and GPU environment. The results show that large models take around 3 seconds for a sentence in a CPU and take around 1 second for a sentence in a GPU on average while the base models take approximately one third of that time in both environments. From these results it is clear that MUDES is capable of predicting toxic spans efficiently in any environment. The full set of results are reported in the Appendix. We used a batch size of one, in order to mimic the real world scenario. The full specifications of the CPU and GPU environments are listed in the Appendix.


\section{Conclusion}
This paper introduced {\em MUDES:} {\bf Mu}ltilingual {\bf De}tection of Offensive {\bf S}pans. We evaluated MUDES on the recently released SemEval-2021 Toxic Spans Detection dataset. Our results show that MUDES outperforms the strong baselines of the competition. Furthermore, we show that once MUDES is trained on English data using state of the art cross-lingual transformer models, it is capable of detecting offensive spans in other languages. With MUDES, we release a Python library, four pre-trained models and an user interface. We show that MUDES is efficient to use in real time scenarios even in a non GPU environment.  In future work, we would like to further evaluate MUDES on other datasets. Finally, we would like to implement a flexible multitask architecture capable of detecting offense at both span and post level.

\section*{Acknowledgments}

We would like to thank the SemEval-2021 Toxic
Spans Detection shared task organisers for making this interesting dataset available. We further thank the anonymous reviewers for their insightful feedback.

\bibliography{anthology,custom}
\bibliographystyle{acl_natbib}

\newpage

\section*{Appendix}

\begin{enumerate}[i]
  \item \textbf{Training Configurations}
We used an Nvidia Tesla K80 GPU to train the models. We divided the dataset into a training set and a validation set using 0.8:0.2 split. We fine tuned the learning rate and number of epochs of the model manually to obtain the best results for the validation set. We obtained $1e^-5$ as the best value for learning rate and 3 as the best value for number of epochs for all the languages.  We performed \textit{early stopping} if the validation loss did not improve over 10 evaluation steps. Training large models took around 30 minutes while training base models took around 10 minutes. In addition to the learning rate and number of epochs we used the parameter values mentioned in Table \ref{tab:parameter}. We kept these values as constants. 

\begin{table}[!ht]
\centering
\setlength{\tabcolsep}{4.5pt}
\scalebox{0.85}{
\begin{tabular}{ll}
\hline
\bf Parameter & \bf Value  \\ \hline
adam epsilon & 1e-8       \\
warmup ratio & 0.1    \\
warmup steps  & 0       \\
max grad norm & 1.0        \\
max seq. length & 140        \\
gradient accumulation steps & 1 \\
 \hline
\end{tabular}
}
\caption{Parameter Specifications.}
\label{tab:parameter}
\end{table}
  \item \textbf{Hardware Specifications}

In Table \ref{tab:gpu} and in Table \ref{tab:cpu} we mention the specifications of the GPU and CPU we used for the experiments of the paper. For the training of the MUDES models, we mainly used the GPU. For the efficiency experiments mentioned in Section \ref{sec:eff} we used both GPU and CPU environments.

\begin{table}[!ht]
\centering
\setlength{\tabcolsep}{4.5pt}
\scalebox{0.85}{
\begin{tabular}{ll}
\hline
\bf Parameter & \bf Value  \\ \hline
GPU & Nvidia K80       \\
GPU Memory & 12GB    \\
GPU Memory Clock  & 0.82GHz       \\
Performance & 4.1 TFLOPS        \\
No. CPU Cores & 2        \\
RAM & 12GB        \\
 \hline
\end{tabular}
}
\caption{GPU Specifications.}
\label{tab:gpu}
\end{table}

\begin{table}[!ht]
\centering
\setlength{\tabcolsep}{4.5pt}
\scalebox{0.85}{
\begin{tabular}{ll}
\hline
\bf Parameter & \bf Value  \\ \hline
CPU Model Name & Intel(R) Xeon(R)       \\
CPU Freq. & 2.30GHz    \\
No. CPU Cores  & 2      \\
CPU Family & Haswell        \\
RAM & 12GB        \\
 \hline
\end{tabular}
}
\caption{CPU Specifications.}
\label{tab:cpu}
\end{table}

\item \textbf{Run time}

As expected base models perform efficiently than the large models in both environments. Large models take around 3 seconds for a sentence in a CPU and take around 1 second for a sentence in a GPU while the base models take approximately one third of that time in both environments. From these results it is clear that MUDES is capable of predicting toxic spans efficiently in any environment. 

\begin{table}[!ht]
\centering
\setlength{\tabcolsep}{4.5pt}
\scalebox{0.85}{
\begin{tabular}{llcc}
\hline
\bf Model & \bf GPU Time   & \bf CPU Time &  \\ \hline
en-base & 35.51 & 100.81       \\
en-large & 100.36 & 315.72   \\
multilingual-base  & 36.23 & 115.98      \\
multilingual-large & 120.54 & 335.65       \\
 \hline
\end{tabular}
}
\caption{Time taken to do predictions on 100 sentences in seconds.}
\label{tab:time}
\end{table}

\end{enumerate}

\section*{Ethics Statement}

MUDES is essentially a web-based visualization tool with predictive models trained on multiple publicly available datasets. The authors of this paper used datasets referenced in this paper which were previously collected and annotated. No new data collection has been carried out as part of this work. 
We have not collected or processed writers'/users' information nor have we carried out any form of user profiling protecting users' privacy and identity. 

We understand that every dataset is subject to intrinsic bias and that computational models will inevitably learn biased information from any dataset. We believe that MUDES will help coping with biases in datasets and models as it features: (1) a freely available Python library that other researchers can use to train new models on other datasets; (2) a web-based visualizing tool that can help efforts in reducing biases in offensive language identification as they can be used to process and visualize potentially offensive spans new data. Finally, unlike models trained at the post level, the projected annotation of spans allows users to understand which part of the instance is considered offensive by the models. 

\end{document}